\pdfoutput=1
\documentclass[11pt,a4paper]{article}
\usepackage{arabtex}
\usepackage{utf8}
\usepackage[hyperref]{eacl2021}
\usepackage{times}
\usepackage{multirow}
\usepackage{latexsym}

\usepackage{microtype}

\aclfinalcopy 

\title{Dialect Identification in Nuanced Arabic Tweets Using Farasa Segmentation and AraBERT}

\author{Anshul Wadhawan \\
  Flipkart Private Limited \\
  \texttt{anshul.wadhwan@flipkart.com} } 

\date{}

\begin{document}
\setcode{utf8}
\maketitle
\begin{abstract}

This paper presents our approach to address the EACL WANLP-2021 Shared Task~1: Nuanced Arabic Dialect Identification (NADI). The task is aimed at developing a system that identifies the geographical location(country/province) from where an Arabic tweet in the form of modern standard Arabic or dialect comes from. We solve the task in two parts. The first part involves pre-processing the provided dataset by cleaning, adding and segmenting various parts of the text. This is followed by carrying out experiments with different versions of two Transformer based models, AraBERT and AraELECTRA. Our final approach achieved macro F1-scores of 0.216, 0.235, 0.054, and 0.043 in the four subtasks, and we were ranked second in MSA identification subtasks and fourth in DA identification subtasks.

\end{abstract}

\section{Introduction}

Spoken by about 500 million people around the world, Arabic is the biggest part of the Semitic language family. Being the official language of almost 22 countries belonging to the Middle-East North Africa (MENA) region, it is not only an integral member of the six official UN languages, but also fourth most used language on the Internet \cite{GUELLIL2019}. Middle East contributes to 164 million internet users and North Africa contributes to 121 million internet users. Comparing with other languages, Arabic language has received little attention in modern computational linguistics, despite its religious, political and cultural significance. However, with rapid development of tools and techniques delivering state-of-the-art performance in many language processing tasks, this negligence is being taken care of.     

The presence of various dialects and complex morphology are some of the distinguishing factors prominent in the Arabic language. Also, the informal nature of conversations on social media and the differences in Modern Standard Arabic (MSA) and Dialectical Arabic (DA), both significantly increase this complexity. While DA is used for informal daily communication, MSA is used for formal writing. Social media is the home for both of these forms, with the former being the most common form. Lack of data is the primary reason why many of the Arabic dialects remain understudied. With the availability of diverse data belonging to 21 Arab countries, this bottleneck can be diminished. The Nuanced Arabic Dialect Identification (NADI), with this goal, is the task of automatic detection of the source variety of a given text or speech segment. 

Previously, on the lines of Arabic dialect identification, there have been approaches focusing on coarse-grained regional varieties such as Levantine or Gulf \cite{elaraby-abdul-mageed-2018-deep,zaidan,elfardy-diab-2013-sentence} or country level varieties \cite{bouamor2019madar,zhang-abdul-mageed-2019-army}. There have been tasks that involved city level classification on human translated data \cite{salameh2018fine}. Some tasks have focused on country and province level classification simultaneously \cite{mageed:2020:nadi}.

In this paper, we present our process to tackle the WANLP-2021 Shared Task 1. The paper is organised in the following way: Section 2 presents the problem statement and details of the provided dataset. Section 3 describes a modularised process that we inculcate as part of methodology. Section 4 describes the experiments that were conducted, with detailed statistics about the dataset, system settings and results of these experiments. A brief conclusion of the paper with the potential prospects of our study are presented in Section 5.

\section{Task Definition}

The WANLP-2021 Shared Task 1 \cite{mageed:2021:nadi} is based on a multi-class classification problem where the aim is to recognize which country or province an Arabic tweet in the form of modern standard Arabic or dialect belongs to. The task targets dialects at the province-level, and also focuses on naturally-occurring fine-grained dialects at the sub-country level. The NADI 2021 task promotes efforts made towards distinguishing both modern standard Arabic (MSA) and dialects (DA) according to their geographical origin, focusing on fine-grained dialects with new datasets. The provided data comes from the domain of Twitter and covers 100 provinces from 21 Arab countries. The task is divided into 4 subtasks as described below:

Subtask 1.1: Country-level MSA identification

Subtask 1.2: Country-level DA identification

Subtask 2.1: Province-level MSA identification

Subtask 2.2: Province-level DA identification


The training dataset has a total of 21,000 tweet, validation and test datasets have 5,000 tweets each. Every example belongs to one of 100 provinces of 21 Arab countries. Additional 10M unlabeled tweets are provided that can be used in developing the systems for either or both of the tasks. F-score, Accuracy, Precision and Recall are the evaluation metrics. However, the official metric of evaluation is the Macro Averaged F-score.

\section{Methodology}

We present our methodology in two parts. The first part in the methodology is data preprocessing. This is followed by experimenting with different transformer based models for the task at hand. Both these parts have been described in detail in the following sub sections.

\subsection{Data Pre-Processing}

Transformer based models, that we plan to fine tune on our dataset, are pre-trained on processed rather than raw data. Owing to the variations in expression of opinions among users belonging to different parts of the world, the tweets fetched from the website are a clear representation of these variations. We find these variations on randomly checking the given examples in different forms. It is common for users to use slang words on the Twitter platform, and post non-ascii characters like emojis. Also, spelling errors, user mentions and URLs are prominent in tweets of most users. These parts within the tweets do not contribute to being informative towards deciding the geographical location of the tweet as they correspond to noise. Thus, the given dataset is cleaned in the following ways, so that the data used for fine tuning has a similar distribution to that used for the pre-training process: 

\begin{enumerate}
\item Perform Farasa segmentation (for select models only) \cite{abdelali-etal-2016-farasa}.
\item Replace all URLs with [ \< رابط > ], emails with [~\<بريد > ], mentions with [ \< مستخدم > ].
\item Remove HTML line breaks and markup, unwanted characters like emoticons, repeated characters ($>$ 2) and extra spaces.
\item Insert whitespace before and after all non Arabic digits or English Digits and Alphabet and the 2 brackets, and between words and numbers or numbers and words.

\end{enumerate}

\subsection{Transformer Based Models}

The domains of speech recognition \cite{graves} and computer vision \cite{krizhevsky} have largely utilised different deep learning techniques and produced significant improvements over the traditional machine learning techniques. In the domain of natural language processing, most deep learning based techniques until now utilised word vector representations \cite{bengio,yih2011,mikolov} for different classification tasks. Lately, transformer based approaches have shown significant progress towards many NLP benchmarks \cite{vaswani2017attention}, including text classification \cite{chang2020taming}, owing to their ability to build proficient language models. As an output of the pre-training process, embeddings are produced which are utilised for finer tasks. 

\begin{table*}[]
\begin{tabular}{|l|l|l|l|l|l|l|}
\hline
\multicolumn{1}{|c|}{\multirow{2}{*}{\textbf{Model}}} & \multicolumn{2}{c|}{\textbf{Size}}                                      & \multicolumn{1}{c|}{\multirow{2}{*}{\textbf{Pre-Segmentation}}} & \multicolumn{3}{c|}{\textbf{Dataset}}                                                                                  \\ \cline{2-3} \cline{5-7} 
\multicolumn{1}{|c|}{}                                & \multicolumn{1}{c|}{\textbf{MB}} & \multicolumn{1}{c|}{\textbf{Params}} & \multicolumn{1}{c|}{}                                           & \multicolumn{1}{c|}{\textbf{\#Sentences}} & \multicolumn{1}{c|}{\textbf{Size}} & \multicolumn{1}{c|}{\textbf{\#Words}} \\ \hline
AraBERTv0.2-base                                      & 543MB                            & 136M                                 & No                                                              & 200M                                      & 77GB                               & 8.6B                                  \\ \hline
AraBERTv0.2-large                                     & 1.38G                            & 371M                                 & No                                                              & 200M                                      & 77GB                               & 8.6B                                  \\ \hline
AraBERTv2-base                                        & 543MB                            & 136M                                 & Yes                                                             & 200M                                      & 77GB                               & 8.6B                                  \\ \hline
AraBERTv2-large                                       & 1.38G                            & 371M                                 & Yes                                                             & 200M                                      & 77GB                               & 8.6B                                  \\ \hline
AraBERTv0.1-base                                      & 543MB                            & 136M                                 & No                                                              & 77M                                       & 23GB                               & 2.7B                                  \\ \hline
AraBERTv1-base                                        & 543MB                            & 136M                                 & Yes                                                             & 77M                                       & 23GB                               & 2.7B                                  \\ \hline
\end{tabular}
\caption{Model Pre-training Parameters}
\label{tab:my-table}
\end{table*}

\subsubsection{AraBERT} AraBERT is an Arabic pretrained language model based on Google's BERT architecture \cite{antoun2020arabert}. There are six versions of the model: AraBERTv0.1-base, AraBERTv0.2-base, AraBERTv0.2-large, AraBERTv1-base, AraBERTv2-base and AraBERTv2-large. For these variations, the model parameters with respect to the pre-training process have been depicted in Table 1. 

\subsubsection{AraELECTRA} Being a method for self-supervised language representation learning, ELECTRA has the ability of making use of lesser computations for the task of pre-training transformers \cite{antoun2020araelectra}. Similar to the objective of discriminator of a Generative Adversarial Network, ELECTRA models are trained with the goal of distinguishing fake input tokens from the real ones. On the Arabic QA dataset, AraELECTRA achieves state-of-the-art results. 

For all new AraBERT and AraELECTRA models, the same pretraining data is used. The dataset that is used for pre-training, before the application of Farasa Segmentation, has a total of 82,232,988,358 characters or 8,655,948,860 words or 200,095,961 lines, and has a size of 77GB. Initially, several websites like OSCAR unshuffled and filtered, Assafir news articles, Arabic Wikipedia dump from 2020/09/01, The OSIAN Corpus and The 1.5B words Arabic Corpus, were crawled to create the pre-training dataset. Later, unshuffled OSCAR corpus, after thorough filtering, was added to the previous dataset used in AraBERTv1 without including the data from the above mentioned crawled websites to create the new dataset. 

\begin{table}[]
\centering
\begin{tabular}{|l|l|l|l|l|}
\hline
\multicolumn{1}{|c|}{\multirow{2}{*}{\textbf{Country}}} & \multicolumn{2}{c|}{\textbf{DA}}                                        & \multicolumn{2}{c|}{\textbf{MSA}}                                       \\ \cline{2-5} 
\multicolumn{1}{|c|}{}                                  & \multicolumn{1}{c|}{\textbf{Train}} & \multicolumn{1}{c|}{\textbf{Dev}} & \multicolumn{1}{c|}{\textbf{Train}} & \multicolumn{1}{c|}{\textbf{Dev}} \\ \hline
Algeria                                                 & 1809                                & 430                               & 1899                                & 427                               \\ \hline
Bahrain                                                 & 215                                 & 52                                & 211                                 & 51                                \\ \hline
Djibouti                                                & 215                                 & 27                                & 211                                 & 52                                \\ \hline
Egypt                                                   & 4283                                & 1041                              & 4220                                & 1032                              \\ \hline
Iraq                                                    & 2729                                & 664                               & 2719                                & 671                               \\ \hline
Jordan                                                  & 429                                 & 104                               & 422                                 & 103                               \\ \hline
Kuwait                                                  & 429                                 & 105                               & 422                                 & 103                               \\ \hline
Lebanon                                                 & 644                                 & 157                               & 633                                 & 155                               \\ \hline
Libya                                                   & 1286                                & 314                               & 1266                                & 310                               \\ \hline
Mauritania                                              & 215                                 & 53                                & 211                                 & 52                                \\ \hline
Morocco                                                 & 858                                 & 207                               & 844                                 & 207                               \\ \hline
Oman                                                    & 1501                                & 355                               & 1477                                & 341                               \\ \hline
Palestine                                               & 428                                 & 104                               & 422                                 & 102                               \\ \hline
Qatar                                                   & 215                                 & 52                                & 211                                 & 52                                \\ \hline
Saudi\_Arabia                                           & 2140                                & 520                               & 2110                                & 510                               \\ \hline
Somalia                                                 & 172                                 & 49                                & 346                                 & 63                                \\ \hline
Sudan                                                   & 215                                 & 53                                & 211                                 & 48                                \\ \hline
Syria                                                   & 1287                                & 278                               & 1266                                & 309                               \\ \hline
Tunisia                                                 & 859                                 & 173                               & 844                                 & 170                               \\ \hline
UAE                                  & 642                                 & 157                               & 633                                 & 154                               \\ \hline
Yemen                                                   & 429                                 & 105                               & 422                                 & 88                                \\ \hline
\end{tabular}
\caption{Country Level Data Distribution}
\label{tab:my-table}
\end{table}

\begin{table}[]
\centering
\begin{tabular}{|l|l|}
\hline
\textbf{Parameter}        & \textbf{Value}  \\ \hline
Learning Rate             & 1e-5            \\ \hline
Epsilon (Adam optimizer)  & 1e-8            \\ \hline
Maximum Sequence Length   & 256             \\ \hline
Batch Size (for base models)                & 40               \\ \hline
Batch Size (for large models)                & 4               \\ \hline
\#Epochs                  & 5               \\ \hline
\end{tabular}
\caption{Parameter Values}
\label{tab:my-table}
\end{table}

\begin{table*}[]
\centering
\begin{tabular}{|l|l|l|l|l|l|l|l|l|}
\hline
\multicolumn{1}{|c|}{\multirow{2}{*}{\textbf{Model}}} & \multicolumn{2}{c|}{\textbf{Subtask 1.1}}                                 & \multicolumn{2}{c|}{\textbf{Subtask 1.2}}                                 & \multicolumn{2}{c|}{\textbf{Subtask 2.1}}                                 & \multicolumn{2}{c|}{\textbf{Subtask 2.2}}                                 \\ \cline{2-9} 
\multicolumn{1}{|c|}{}                                & \multicolumn{1}{c|}{\textbf{F1}} & \multicolumn{1}{c|}{\textbf{A}} & \multicolumn{1}{c|}{\textbf{F1}} & \multicolumn{1}{c|}{\textbf{A}} & \multicolumn{1}{c|}{\textbf{F1}} & \multicolumn{1}{c|}{\textbf{A}} & \multicolumn{1}{c|}{\textbf{F1}} & \multicolumn{1}{c|}{\textbf{A}} \\ \hline
AraBERTv0.1-base                                      & 0.283                            & 0.324                           & 0.338                            & 0.390                           & 0.024                            & 0.028                           & 0.025                            & 0.037                           \\ \hline
AraBERTv0.2-base                                      & 0.300                            & 0.344                           & 0.382                            & 0.427                           & 0.038                            & 0.042                           & 0.035                            & 0.051                           \\ \hline
AraBERTv0.2-large                                     & 0.304                            & 0.343                           & 0.362                            & 0.413                           & 0.022                            & 0.030                           & 0.029                            & 0.041                           \\ \hline
AraBERTv1-base                                        & 0.281                            & 0.318                           & 0.306                            & 0.377                           & 0.032                            & 0.040                           & 0.019                            & 0.033                           \\ \hline
AraBERTv2-base                                        & 0.309                            & 0.347                           & 0.389                            & 0.432                           & 0.029                            & 0.038                           & 0.034                            & 0.048                           \\ \hline
AraBERTv2-large                                       & 0.315                            & 0.346                           & 0.416                            & 0.450                           & 0.001                            & 0.010                           & 0.001                            & 0.010                           \\ \hline
AraELECTRA-base-generator                             & 0.106                            & 0.231                           & 0.165                            & 0.285                           & 0.005                            & 0.018                           & 0.006                            & 0.022                           \\ \hline
AraELECTRA-base-discriminator                         & 0.192                            & 0.281                           & 0.280                            & 0.375                           & 0.007                            & 0.020                           & 0.006                            & 0.026                           \\ \hline
\end{tabular}
\caption{Validation Set Results}
\label{tab:my-table}
\end{table*}

\begin{table}[]
\centering
\begin{tabular}{|c|c|c|c|c|}
\hline
              & \textbf{M-F1} & \textbf{A} & \textbf{P} & \textbf{R}\\ \hline
Subtask 1.1 & 0.216  & 0.317 & 0.321 & 0.189 \\ \hline
Subtask 1.2 & 0.235 & 0.433 & 0.280 & 0.233   \\ \hline
Subtask 2.1 & 0.054 & 0.060 & 0.061 & 0.060   \\ \hline
Subtask 2.2 & 0.043 & 0.053 & 0.044 & 0.051  \\ \hline
\end{tabular}
\caption{Test Set Results}
\label{table:1}
\end{table}

\section{Experiments}

We experiment with eight transformer based models using the given training and validation sets. We calculate the final test predictions by fine tuning the most efficient model, which is decided by the scores produced above, with the concatenated labeled training and validation splits. This is followed by evaluating the test set on this fine tuned model.
This section presents the Country-level dataset distribution, system settings, results of our research followed by a descriptive analysis of our system. 

\subsection{Dataset}

The country-wise distribution of the provided training and validation splits, for both the tasks of MSA and DA, are shown in Table 2.

\subsection{System Settings}

We make use of pre-trained AraBERT and AraELECTRA models, with the names of bert-base-arabert, bert-base-arabertv01, bert-large-arabertv2, bert-base-arabertv2, bert-large-arabertv02, bert-base-arabertv02, araelectra-base-generator and araelectra-base-discriminator for fine-tuning the transformer based models. We use hugging-face\footnote{\url{https://huggingface.co/transformers/}} API to fetch the pre-trained transformer based models, and then fine tuned the same on our dataset. The hyper parameters used for fine tuning these models have been specified in Table 3.

\subsection{Results and Analysis}

For all subtasks, the performance results of proposed models on the provided validation set with reference to accuracy(A) and weighted F1 scores(F1) are shown in Table 4.

From Table 4, we conclude that:
\begin{enumerate}
\item For most of the subtasks, one of the base models performs almost as good as the best performing large model.
\item AraELECTRA models seem to perform worse than all AraBERT models, possibly due to their specialization in handling GAN related tasks, which are different from classification based tasks.
\item AraBERTv2-large out performs all other models for subtasks 1.1 and 1.2. For subtasks 2.1 and 2.2, AraBERTv0.2-base produces the best results on the validation set.
\end{enumerate}

From the above results, we choose AraBERTv2-large for subtasks 1.1, 1.2 and AraBERTv0.2-base for subtasks 2.1, 2.2 to be the primary models to fine tune on the concatenated training and validation set as well as carry out inferences on the unseen dataset. The final test set results in terms of Macro F1 Score(M-F1), Recall(R), Accuracy(A) and Precision(P) are specified in Table 5.

\section{Conclusion and Future Work}

In this paper, we present a comprehensive overview of the approach that we employed to solve the EACL WANLP-2021 Shared Task 1. We tackle the given problem in two parts. The first part involves pre processing the given data by modifying various parts of the text. The second part involves experimenting with different versions of two Transformer based networks, AraBERT and AraELECTRA, all pre-trained on Arabic text. Our final submissions for the four subtasks are based on the best performing version of AraBERT model. With Macro Averaged F1-Score as the final evaluation criteria, our approach fetches a private leaderboard rank of 2 for MSA identification and 4 for DA identification. In the future, we aim to utilise other features relevant for classification tasks like URLs, emoticons, and experiment with ensembles of transformer based and word vector based input representations. 

\bibliography{eacl2021}

\begin{thebibliography}{19}
\expandafter\ifx\csname natexlab\endcsname\relax\def\natexlab#1{#1}\fi

\bibitem[{Abdelali et~al.(2016)Abdelali, Darwish, Durrani, and
  Mubarak}]{abdelali-etal-2016-farasa}
Ahmed Abdelali, Kareem Darwish, Nadir Durrani, and Hamdy Mubarak. 2016.
\newblock \href {https://doi.org/10.18653/v1/N16-3003} {{F}arasa: A fast and
  furious segmenter for {A}rabic}.
\newblock In \emph{Proceedings of the 2016 Conference of the North {A}merican
  Chapter of the Association for Computational Linguistics: Demonstrations},
  pages 11--16, San Diego, California. Association for Computational
  Linguistics.

\bibitem[{Abdul-Mageed et~al.(2020)Abdul-Mageed, Zhang, Bouamor, and
  Habash}]{mageed:2020:nadi}
Muhammad Abdul-Mageed, Chiyu Zhang, Houda Bouamor, and Nizar Habash. 2020.
\newblock {NADI 2020: The First Nuanced Arabic Dialect Identification Shared
  Task}.
\newblock In \emph{Proceedings of the Fifth Arabic Natural Language Processing
  Workshop (WANLP 2020)}, Barcelona, Spain.

\bibitem[{Abdul-Mageed et~al.(2021)Abdul-Mageed, Zhang, Elmadany, Bouamor, and
  Habash}]{mageed:2021:nadi}
Muhammad Abdul-Mageed, Chiyu Zhang, AbdelRahim Elmadany, Houda Bouamor, and
  Nizar Habash. 2021.
\newblock {NADI 2021: The Second Nuanced Arabic Dialect Identification Shared
  Task}.
\newblock In \emph{Proceedings of the Sixth {A}rabic Natural Language
  Processing Workshop (WANLP 2021)}.

\bibitem[{Antoun et~al.()Antoun, Baly, and Hajj}]{antoun2020arabert}
Wissam Antoun, Fady Baly, and Hazem Hajj.
\newblock Arabert: Transformer-based model for arabic language understanding.
\newblock In \emph{LREC 2020 Workshop Language Resources and Evaluation
  Conference 11--16 May 2020}, page~9.

\bibitem[{Antoun et~al.(2020)Antoun, Baly, and Hajj}]{antoun2020araelectra}
Wissam Antoun, Fady Baly, and Hazem Hajj. 2020.
\newblock \href {http://arxiv.org/abs/2012.15516} {Araelectra: Pre-training
  text discriminators for arabic language understanding}.

\bibitem[{Bengio et~al.(2003)Bengio, Ducharme, Vincent, and Janvin}]{bengio}
Yoshua Bengio, R\'{e}jean Ducharme, Pascal Vincent, and Christian Janvin. 2003.
\newblock A neural probabilistic language model.
\newblock \emph{J. Mach. Learn. Res.}, 3(null):1137–1155.

\bibitem[{Bouamor et~al.(2019)Bouamor, Hassan, and Habash}]{bouamor2019madar}
Houda Bouamor, Sabit Hassan, and Nizar Habash. 2019.
\newblock The madar shared task on arabic fine-grained dialect identification.
\newblock In \emph{Proceedings of the Fourth Arabic Natural Language Processing
  Workshop}, pages 199--207.

\bibitem[{Chang et~al.(2020)Chang, Yu, Zhong, Yang, and
  Dhillon}]{chang2020taming}
Wei-Cheng Chang, Hsiang-Fu Yu, Kai Zhong, Yiming Yang, and Inderjit Dhillon.
  2020.
\newblock \href {http://arxiv.org/abs/1905.02331} {Taming pretrained
  transformers for extreme multi-label text classification}.

\bibitem[{Elaraby and Abdul-Mageed(2018)}]{elaraby-abdul-mageed-2018-deep}
Mohamed Elaraby and Muhammad Abdul-Mageed. 2018.
\newblock \href {https://www.aclweb.org/anthology/W18-3930} {Deep models for
  {A}rabic dialect identification on benchmarked data}.
\newblock In \emph{Proceedings of the Fifth Workshop on {NLP} for Similar
  Languages, Varieties and Dialects ({V}ar{D}ial 2018)}, pages 263--274, Santa
  Fe, New Mexico, USA. Association for Computational Linguistics.

\bibitem[{Elfardy and Diab(2013)}]{elfardy-diab-2013-sentence}
Heba Elfardy and Mona Diab. 2013.
\newblock \href {https://www.aclweb.org/anthology/P13-2081} {Sentence level
  dialect identification in {A}rabic}.
\newblock In \emph{Proceedings of the 51st Annual Meeting of the Association
  for Computational Linguistics (Volume 2: Short Papers)}, pages 456--461,
  Sofia, Bulgaria. Association for Computational Linguistics.

\bibitem[{Graves et~al.(2013)Graves, rahman Mohamed, and Hinton}]{graves}
Alex Graves, Abdel rahman Mohamed, and Geoffrey Hinton. 2013.
\newblock \href {http://arxiv.org/abs/1303.5778} {Speech recognition with deep
  recurrent neural networks}.

\bibitem[{Guellil et~al.(2019)Guellil, Saâdane, Azouaou, Gueni, and
  Nouvel}]{GUELLIL2019}
Imane Guellil, Houda Saâdane, Faical Azouaou, Billel Gueni, and Damien Nouvel.
  2019.
\newblock \href {https://doi.org/https://doi.org/10.1016/j.jksuci.2019.02.006}
  {Arabic natural language processing: An overview}.
\newblock \emph{Journal of King Saud University - Computer and Information
  Sciences}.

\bibitem[{Krizhevsky et~al.(2012)Krizhevsky, Sutskever, and
  Hinton}]{krizhevsky}
Alex Krizhevsky, Ilya Sutskever, and Geoffrey~E. Hinton. 2012.
\newblock Imagenet classification with deep convolutional neural networks.
\newblock In \emph{Proceedings of the 25th International Conference on Neural
  Information Processing Systems - Volume 1}, NIPS'12, page 1097–1105, Red
  Hook, NY, USA. Curran Associates Inc.

\bibitem[{Mikolov et~al.(2013)Mikolov, Sutskever, Chen, Corrado, and
  Dean}]{mikolov}
Tomas Mikolov, Ilya Sutskever, Kai Chen, Greg Corrado, and Jeffrey Dean. 2013.
\newblock Distributed representations of words and phrases and their
  compositionality.
\newblock In \emph{Proceedings of the 26th International Conference on Neural
  Information Processing Systems - Volume 2}, NIPS'13, page 3111–3119, Red
  Hook, NY, USA. Curran Associates Inc.

\bibitem[{Salameh et~al.(2018)Salameh, Bouamor, and Habash}]{salameh2018fine}
Mohammad Salameh, Houda Bouamor, and Nizar Habash. 2018.
\newblock Fine-grained arabic dialect identification.
\newblock In \emph{Proceedings of the 27th International Conference on
  Computational Linguistics}, pages 1332--1344.

\bibitem[{Vaswani et~al.(2017)Vaswani, Shazeer, Parmar, Uszkoreit, Jones,
  Gomez, Kaiser, and Polosukhin}]{vaswani2017attention}
Ashish Vaswani, Noam Shazeer, Niki Parmar, Jakob Uszkoreit, Llion Jones,
  Aidan~N. Gomez, Lukasz Kaiser, and Illia Polosukhin. 2017.
\newblock \href {http://arxiv.org/abs/1706.03762} {Attention is all you need}.

\bibitem[{Yih et~al.(2011)Yih, Toutanova, Platt, and Meek}]{yih2011}
Wen-tau Yih, Kristina Toutanova, John~C. Platt, and Christopher Meek. 2011.
\newblock \href {https://www.aclweb.org/anthology/W11-0329} {Learning
  discriminative projections for text similarity measures}.
\newblock In \emph{Proceedings of the Fifteenth Conference on Computational
  Natural Language Learning}, pages 247--256, Portland, Oregon, USA.
  Association for Computational Linguistics.

\bibitem[{Zaidan and Callison-Burch(2014)}]{zaidan}
Omar~F. Zaidan and Chris Callison-Burch. 2014.
\newblock \href {https://doi.org/10.1162/COLI\_a\_00169} {Arabic dialect
  identification}.
\newblock \emph{Computational Linguistics}, 40(1):171--202.

\bibitem[{Zhang and Abdul-Mageed(2019)}]{zhang-abdul-mageed-2019-army}
Chiyu Zhang and Muhammad Abdul-Mageed. 2019.
\newblock \href {https://doi.org/10.18653/v1/W19-4637} {No army, no navy:
  {BERT} semi-supervised learning of {A}rabic dialects}.
\newblock In \emph{Proceedings of the Fourth Arabic Natural Language Processing
  Workshop}, pages 279--284, Florence, Italy. Association for Computational
  Linguistics.

\end{thebibliography}
\bibliographystyle{acl_natbib}

\end{document}